# Building another Spanish dictionary, this time with GPT-4

Miguel Ortega-Martín,[1,2] Óscar García-Sierra,[1,2] Alfonso Ardoiz,[1,2]
Juan Carlos Armenteros,[1] Ignacio Garrido,[1] Jorge Álvarez,[1]
Camilo Torrón,[1] Iñigo Galdeano,[1] Ignacio Arranz,[1]
Oleg Vorontsov,[1] Adrián Alonso,[1,3]

[1]dezzai
[2]Universidad Complutense de Madrid
[3]Universidad Rey Juan Carlos

{m.ortega, oscar.garcia, alfonso.ardoiz, juancarlos.armenteros,
ignacio.garrido, jorge.alvarez, camilo.torron, i.galdeano,
ignacio.arranz, o.vorontsov, a.alonso}@dezzai.com

**Abstract:** We present the "Spanish Built Factual Freectianary 2.0" (Spanish-BFF-2) as the second iteration of an AI-generated Spanish dictionary. Previously, we developed the inaugural version of this unique free dictionary employing GPT-3. In this study, we aim to improve the dictionary by using GPT-4-turbo instead. Furthermore, we explore improvements made to the initial version and compare the performance of both models.
**Keywords:** Large Language Models, Computational lexicography, Dictionaries, Generative AI

## 1 Introduction

Dictionaries have been extensively employed as foundational linguistic references since their inception. Traditionally arranged alphabetically, these linguistic compendiums exist in diverse manifestations, ranging from monolingual and bilingual to general and domain-specific. Each is characterized by unique features such as structure and content.

Significant advancements have been achieved in probing and undercutting space limitations through digitization. This work aspires to contribute to assembling new dictionaries straightforwardly and improve users' search experience by equipping them with more flexibility and openness.

The intrinsic dynamics of Large Language Models (LLMs) and their generative capabilities imply an inclusive approach to meaning. Meaning is the aggregate of occurrences cognate with a lexical unit derived from the eclectic sources utilized during the model's training. Consequently, the outcome is a statistical representation aligning with its prevailing understanding.

In our earlier endeavor, we pioneered the first Spanish monolingual dictionary generated by artificial intelligence (Ortega-Martín et al., 2023). This initial proposal comprised 66,353 lemmas and excluded polysemy and part-of-speech (POS) tags. Our current effort aspires to enhance our methodology by covering all Spanish lemmas without exclusions and providing illustrative usage examples. It is imperative to acknowledge that despite our efforts to bestow information for every one of the 94,472 Spanish lemmas, approximately 17,379 failed. Consequently, these instances are excluded from the final dictionary but are subject to investigation in the error analysis section. Ultimately, the definitive dictionary contains 77,093 words.

This paper is organized as follows: Section 2 focuses on the realm of LLMs, while Section 3 provides an overview of lexicography. Section 4 outlines our experimental setup, followed by a thorough examination of the generated dictionary in Section 5. Section 6 elucidates errors, leading to the conclusive findings in Section 7. An exploration of the inherent limitations of our work and future work is presented in Section 8, and ethical considerations are addressed in the Ethics Statement (Section 9).

The contributions of this paper are summarized as follows:

- We have constructed, to our understanding, the second freely accessible AI-generated dictionary. Nevertheless, it is the first of its kind in Spanish. Mainly, this dictionary encompasses Spanish lemmas, definitions, and examples contributed utilizing GPT-4-turbo. It also constitutes a more intricate and comprehensive version than its precursor, generated with GPT-3 (Ortega-Martín et al., 2023).

- We have evaluated the role of both models by conducting a comparative analysis of the two Spanish-BFF versions.

## 2 Large Language Models

LLMs play a crucial role in the field of Natural Language Processing (NLP). Particularly, GPT models (Brown et al., 2020) have garnered recognition as prominent encoder-decoder architectures. These text-to-text models are rooted in vector representations of words or subwords. Recent advancements, exemplified by InstructGPT (Ouyang et al., 2022), set themselves apart by incorporating user intents to enhance functioning. InstructGPT, a fine-tuned version of GPT-3 with 1,3 billion parameters, considers supervised learning and Reinforcement Learning from Human Feedback (RLHF). Another variant, ChatGPT,[1] equipped with 175 million parameters, is designed purposely for user interactions. More recently, GPT-4-turbo[2] has been introduced as an advanced version of GPT-4,[3] boasting 1,76 trillion parameters (Achiam et al., 2023), a 128,000 context window, and a maximum output limit of 4096 tokens. Prompting is emphasized in these models, where a prompt —a piece of text— narrows down the input context, thereby improving the quality of produced text (Ouyang et al., 2022).

## 3 Lexicography

Lexicography is the study of dictionaries and their compilation. It can be categorized into two primary branches: theoretical lexicography, which explores theories concerning the arrangements and contents of dictionaries, and practical lexicography, which concentrates on the tangible creation of dictionaries (Bergenholtz y Gouws, 2012). The meticulous process of constructing a dictionary adheres to detailed guidelines and principles to ensure the efficacy of referencing and comprehension (Jackson, 2013).

Historically, dictionaries have served as the predominant avenue for accessing meaning. Within this context, manifold classifications have been proposed, with one of the most prevalent being the following categorization:

- Alphabetical arrangement: This category encompasses semasiological dictionaries, starting from the signifier or *definiendum*, and facilitating access to the *definiens* or lexicographic definition.

- Systematic arrangement: Within this taxonomy, onomasiological dictionaries enable users to locate the corresponding *definiendum* or identify closely related terms through assorted relationships and established arrangements.

- Ideological and analogical arrangement: Which encloses ideological or onomasiological dictionaries.

The dictionary layout includes three fundamental components: external matter (comprising supplementary resources or usage guidelines), macro-structure, and micro-structure (Kirkness, 2004). Macro-structure involves the organization of lemmas within the dictionary, with variations in size and format dependent on the type and domain of the dictionary (Hausmann y Wiegand, 1989). Micro-structure pertains to the linguistic data encapsulated inside each entry. At the co-re of dictionaries are definitions, conventionally encompassing elements such as a generic term, POS tags indicating the word's class, a listing of senses arranged by specific sorting rules, and occasional usage examples. Moreover, dictionaries may consider linguistic annotations concerning spelling, pronunciation, base and inflected forms, morphological information, and semantic nuances such as synonyms, antonyms, hypernyms, or hyponyms (Kirkness, 2004).

---
[1] https://openai.com/blog/chatgpt/
[2] https://platform.openai.com/docs/models/gpt-4-and-gpt-4-turbo
[3] https://openai.com/gpt-4

Computers play a pivotal role in contemporary lexicography, with electronic storage of textual data in corpora and varied computerized presentation of lexicological work representing significant advancements (Kirkness, 2004). Computational lexicography develops annotated dictionaries and lexicons from extensive raw text volumes. While conventional digital dictionaries demand substantial resources, modern LLMs operate autonomously, assembling vocabularies and optimizing computational expenses. Although this adaptability is advantageous, it may introduce a degree of generalization.

Recent methodologies leverage dictionaries to generate word embeddings (Hill et al., 2016), handling word sets from definitions to compute embeddings (Ortega-Martín, 2021). This approach allows access to previously untapped and unexploited lexicographic information when training models on Internet corpora. Definition Modeling (DM) represents an NLP task that aims to model meanings from word embeddings (Noraset et al., 2017) running Recurrent Neural Networks (RNNs) (Schmidt, 2019) for qualitative and quantitative error analyses of generated definitions. Bevilacqua, Maru, y Navigli (2020) develop contextual glosses from words and phrases using a BART model (Lewis et al., 2019).

Regardless of the widespread use of generative models in diverse NLP tasks, their potential for producing exhaustive dictionaries still needs to be explored. Malkin et al. (2021) uses GPT-3 for supplying meanings to newly coined terms. Our previous endeavor (Ortega-Martín et al., 2023) asserts the creation of the inaugural *freectianary*, encompassing 66,353 Spanish lemmas. However, it is pertinent to remark that the prior work did not account for polysemy, lacked POS tags, and omitted sentence examples for the provided definitions.

## 4 Experimental set-up

We manipulate a curated list comprising 94,472 Spanish lemmas for our approach. It contemplates polysemy and distinct POS tags, which were neglected in our first submission, Spanish-BFF-1 (Ortega-Martín et al., 2023). Besides, in this proposition, we run GPT-4-turbo and request example sentences.

The methodology for rendering definitions requires submitting a prompt-based GPT-4-turbo query to the OpenAI API. We handle batches of 32 explicitly prepared lemmas per query. The prompt specifies the instructions for yielding definitions and examples for each lemma and category. Additionally, it adds a few-shot approach by feeding the model with a couple of examples of the expected output. Building the entire dictionary took around 90 hours. Nonetheless, it should be noted that response speed varies significantly in time according to the demand.

The second iteration of the "Spanish Built Factual Freectianary" (Spanish-BFF-2) is accessible on the Hugging Face hub[4] and on our GitHub repository.[5]

## 5 Results and contrast

To rate this version of the generated dictionary and the improvement in succeeding GPT models, we conduct a qualitative and quantitative analysis of definitions versus previous Spanish-BFF-1 and "Diccionario de la Lengua Española" (DLE).[6]

### 5.1 Qualitative analysis

As occurred with GPT-3 in Spanish-BFF-1 (Ortega-Martín et al., 2023), GPT-4-turbo exhibits commendable lexicographic capabilities in following lexicographic principles. When restricting a noun meaning (see Appendix A.1), GPT-4-turbo adeptly alludes to the word's class, frequently introducing another verb to convey the meaning effectively. Also, for verbs, it uses a verb with a broader meaning (see Appendix A.2). In the case of adjectives (see Appendix A.3), the model commonly alludes to forms such as "que..."("that.."), "relativo a ..."(relative to...") and "se refiere a..."(it refers to...), or mentions synonyms of the defined word.

---

[4] https://huggingface.co/datasets/MMG/spanishBFF2
[5] https://github.com/dezzai/Spanish-BFF-2
[6] https://dle.rae.es/

Adverbs are often described with phrases like "de manera..." ("in a [related adjective] way") or by delivering synonyms (see Appendix A.4). Ultimately, the latter model improves the implementation of its predecessor regarding the restriction to incorporating the defined word in the definition or its vocabulary richness.

### 5.1.1 Example sentences

We explicitly prompted GPT-4-turbo to supply an example sentence for each definition. While a direct comparison with Spanish-BFF-1 is unattainable, carefully investigating these examples is feasible. The generated sentences are concise yet accurate when the corresponding definitions are correct. If GPT-4-turbo is unfamiliar with a particular lemma, it refrains from forging an example. The sentences are generally simple, illustrating various functions of the target word, such as acting as a subject, direct object, or different kinds of complements (see Appendix A.5).

## 5.2 Quantitative analysis

Similarly to Spanish-BFF-1, to benchmark the results, we parse the output of queries for these lemmas against DLE, which aspires to encompass all Spanish words. It is crucial to emphasize that we neither store nor manipulate outputs from DLE for any commercial purposes. This procedure is strictly for research, aiming to discern the performance of the proposed dictionary against a trusted source. Out of 94,472 candidate words, we could parse a gold standard definition for 76,963 of them in DLE, so this section is restricted to those terms.

We divide the quantitative evaluation of the definitions into three approaches. Firstly, we implement cosine similarity for lemmas with just one meaning in both Spanish-BFF-1 and DLE to inspect the definitions' grade (Noraset et al., 2017). In order to calculate cosine similarity, we apply the model named "distiluse-base-multilingual-cased-v2".[7] Secondly, whether monosemous entries correspond to polysemous words in DLE or the other way round, we rank the sense with the highest resemblance of the cosine similarity. Lastly, we execute supervised classification metrics such as precision, recall, and F1 scores to gauge the polysemy gain.

### 5.2.1 Monosemy

There are 73,558 lemmas with a single definition in Spanish-BFF-2, while there are 49,813 such lemmas in DLE. The intersection of these sets amounts to 49,114, as shown in Table 1. Therefore, in line with Table 2, 66,8 % of words with just one meaning in Spanish-BFF-2 are monosemous in DLE, but 98,6 % of monosemous words in DLE have one definition in Spanish-BFF-2. GPT4-turbo seems reluctant to provide further senses in the vast majority of cases.

Table 3 gives monosemy metrics between the dictionaries regarding the DLE ideal. Like GPT-3, GPT-4-turbo performs better in defining adverbs or adjectives than verbs or nouns. On top of that, GPT-4-turbo outperforms GPT-3 across all POS tags except adverbs, which undergo an operation decline. Additionally, the spread increases slightly despite essentially showing a higher average of the cosine similarity.

Acknowledging the high standard of the ground-truth definitions used for comparison is essential. In Spanish-BFF-1, GPT-3 demonstrated shorter entries than the related ones in DLE, as illustrated in Table 4. Undeniably, GPT-4-turbo definitions are more extended and nearly aligned with DLE lengths, as shown in Table 4. However, DLE definitions of adverbs are shorter, like those in BFF-1, while GPT-turbo uses more than twice as many words. As seen in Appendix A.6, it seems that GPT-4-turbo adverbs definitions are too complex.

### 5.2.2 Polysemy disagreement

From the 73,558 lemmas in Spanish-BFF-2 with a single definition, 24,444 of them are polysemous in DLE, according to Table 1. Conversely, as noted in Table 2, only 9,8 % of DLE polysemous words have more than one sense in Spanish-BFF-2. That is 2,706 lemmas out of 27,150 actual polysemous ones. Then, as previously stated, the model consistently tends to provide only one definition for most lemmas undeterred by asking for all the known senses.

We are relatively inexpensive in finding polysemy in monosemous words (699). The error analysis section will address these lemmas with one sense in DLE corresponding to polysemous ones in Spanish-BFF-2 since they could predominantly emerge from hallucinations. Nonetheless, polysemous lemmas need to be more adequately covered.

---

[7] https://huggingface.co/sentence-transformers/distiluse-base-multilingual-cased

|  | **Monosemy** | **Polysemy** | **Total** |
|---|---|---|---|
| **Monosemy** | 49,114 | 699 | 49,813 |
| **Polysemy** | 24,444 | 2,706 | 27,150 |
| **Total** | 73,558 | 3,405 | 76,963 |

Table 1: Confusion matrix (actual/prediction) between Spanish-BFF-2 and DLE.

|  | **Precision** | **Recall** | **F1** |
|---|---|---|---|
| **Monosemy** | 0,668 | 0,986 | 0,798 |
| **Polysemy** | 0,795 | 0,098 | 0,177 |

Table 2: Relation between monosemy and polysemy in Spanish-BFF-2 and DLE.

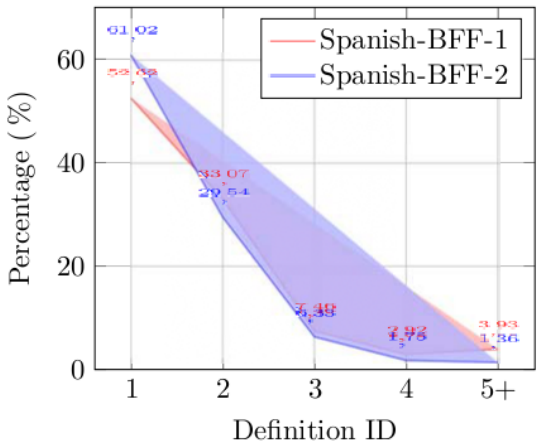

Figure 1: Highest cosine similarity among Spanish-BFF-2 monosemous lemmas and DLE definitions by ID.

As illustrated in Figure 1, the likelihood of deeper definitions in DLE decreases swiftly between versions. This decline corroborates the ranking since both dictionaries are based on the frequency of use. However, some DLE lemmas rely on the chronological order, disrupting the expected hierarchy. Unquestionably, while shallow definitions indicate good health, deeper matches show the tergiversation of the mainstream connotations. Since GPT-4-turbo is sounder than GPT-3 at capturing this statistical trend of meanings, the first entry in the DLE definition fits semantically more agreeably. Even more reliable than for monosemous words, which also occurred in Spanish-BFF-1 (Ortega-Martín et al., 2023), in conformity with Table 3.

Moreover, all the senses supplied for a lemma at DLE have a mean cosine similarity of 0,5535 with the generated one, compared to a score of 0,443 from its predecessor, as shown in Table 5. For all POS tags, cosine similarity when polysemy disagreement occurs is higher than when a lemma has just one sense in both dictionaries, and the tendency is maintained to be higher in adverbs and adjectives than in nouns and verbs.

### 5.2.3 Polysemy

Unlike Spanish-BFF-1, Spanish-BFF-2 comprises 3,405 polysemous words, instead of 27,150 from DLE. Although we explicitly ask GPT to yield all known definitions for a lemma, Table 2 indicates otherwise. The precision achieved for polysemous words is 79 %, yet the recall is remarkably lower. Then, although the model correctly recognizes a polysemous word accurately, it can acknowledge only a small proportion of all polysemous words in the dataset. Classification metrics for monosemous words emphasize the discrepancy, implying that the model is relatively more effective at correctly identifying monosemous words, albeit less precisely than in the case of polysemous words.

The low F1 scores for both categories, 79,8 % for monosemy and 17,7 % for polysemy, reflect the challenge of balancing precision and recall, specifically in pinpointing polysemous words. The significant difference in recall between monosemous and polysemous words suggests that the model struggles to detect the presence of multiple meanings enclosed by a word. This issue could be due to the model's narrow capability to capture and differentiate among the various senses of words in the corpora instructed for training. This investigation underscores the immediate shortcoming of the current submission and the need for improved modeling approaches

|  | Spanish-BFF-1 | | Spanish-BFF-2 | |
|---|---|---|---|---|
| POS tag | Mean | Std Dev | Mean | Std Dev |
| All | 0,3400 | 0,2545 | 0,4422 | 0,2774 |
| Nouns | 0,2922 | 0,2356 | 0,4092 | 0,2770 |
| Adjectives | 0,4943 | 0,2622 | 0,5265 | 0,2766 |
| Verbs | 0,3912 | 0,2346 | 0,4001 | 0,2445 |
| Adverbs | 0,6712 | 0,2329 | 0,5407 | 0,2440 |

Table 3: Mean and standard deviation of the cosine similarity for lemmas with one sense in DLE and Spanish-BFF versions.

|  |  | DLE | | Spanish-BFF-1 | | Spanish-BFF-2 | |
|---|---|---|---|---|---|---|---|
| POS tag | Measure | words | characters | words | characters | words | characters |
| Total | Mean | 10,04 | 58,90 | 8,31 | 49,13 | 10,24 | 62,01 |
|  | Std Dev | 8,38 | 48,19 | 5,01 | 28,27 | 4,93 | 30,01 |
| Nouns | Mean | 10,99 | 64,26 | 9,73 | 57,10 | 11,19 | 68,21 |
|  | Std Dev | 9,46 | 54,76 | 5,00 | 28,31 | 5,28 | 32,19 |
| Adjectives | Mean | 9,04 | 53,35 | 7,23 | 42,48 | 9,26 | 55,03 |
|  | Std Dev | 7,18 | 40,96 | 3,92 | 22,25 | 4,32 | 25,99 |
| Verbs | Mean | 9,00 | 52,88 | 4,37 | 27,12 | 8,72 | 52,56 |
|  | Std Dev | 5,64 | 31,19 | 2,60 | 14,14 | 3,75 | 22,12 |
| Adverbs | Mean | 4,72 | 29,55 | 3,29 | 21,62 | 7,33 | 45,12 |
|  | Std Dev | 4,24 | 25,35 | 1,92 | 10,46 | 3,20 | 18,77 |

Table 4: Statistical distribution (mean and standard deviation) of the definitions' lengths.

to address polysemy in NLP more effectively.

## 6 Error analysis

The primary concern determined in GPT-4-turbo within the framework of Spanish-BFF-2 revolves around undefined terms. Roughly 15 % of the vocabulary in DLE requires an analogous definition in GPT-4-turbo. At first glance, this disparity can be attributed to these words' unconventional and infrequent nature (refer to Appendix B.1). Anyhow, GPT-4-turbo tends to abstain from generating inaccurate definitions.

This idiosyncrasy proves advantageous, departing from the prevalent hallucinations in the LLMs' outputs (Tonmoy et al., 2024). Indeed, detecting hallucinations is an intricate process. Then, we propose filtering words with a generated definition with less than 0.1 cosine similarity to the gold standard as an approach. However, plenty of properly defined words would still refer to unfrequent terms with mere synonyms in DLE. In contrast, GPT-4-turbo provides a complete definition (see Appendix B.2).

We manually examined hallucinations in the remaining words with low cosine similarity between actual and predicted definitions. Four prominent cases were identified.

1. In Appendix B.3, instances occur where a conceivable affix is encountered. The model interprets the whole word as derived from a part of the word that is a known lemma, as exemplified by "ollera", understood as "olla"(meaning) + "era"(meaning), instead of a bird.

2. The subsequent category of hallucinations, once highlighted in Spanish-BFF-1 (Ortega-Martín et al., 2023), entangles GPT-4-turbo occasionally defining a word as a closely spelled, yet distinct, word (see Appendix B.4). Such as "destace", wrongly defined as "destaque". This phenomenon was attributed in Spanish-BFF-1 to the limitations of subword tokenization. An alternative explanation would indicate that LLMs are akin, and the model may interpret unfamiliar terms as erroneous versions

|  | Spanish-BFF-1 | | Spanish-BFF-2 | |
|---|---|---|---|---|
| POS tag | Mean | Std Dev | Mean | Std Dev |
| All | 0,4471 | 0,2143 | 0,5535 | 0,2247 |
| Nouns | 0,4105 | 0,2071 | 0,5097 | 0,2248 |
| Adjectives | 0,5293 | 0,2108 | 0,6237 | 0,2166 |
| Verbs | 0,5020 | 0,2030 | 0,5509 | 0,2038 |
| Adverbs | 0,7028 | 0,2052 | 0,6442 | 0,1944 |

Table 5: Mean and standard deviation of the cosine similarity for lemmas with one sense and BFF and more than one in DLE.

of words resembling them due to likely spelling errors.

3. The model occasionally does not perform well in classifying common names as proper nouns. In principle, they would never appear in a dictionary. Appendix B.5) includes some examples of this oversight.

4. Lastly, as pointed out in the polysemy disagreement section, Appendix B.6 discloses examples in which some predicted senses are fundamentally the same ("asaltador"), we incur in participle definitions against the lexicography principles ("atropellado"), or there exist fabricated definitions ("baboseo").

For all these three types of hallucinations, example sentences are also made up following the sense of the crafted definition (see Appendices B.3 and B.4). It is pivotal to convey the nuance of the situation: the example proposed aligns logically and is internally consistent with the narrative context, even plausible and persuasive, which lends it a convincing air, making it a compelling representative of a common misunderstanding or inventive interpretation. Still, no hallucinations have been detected in the example sentence when the word is accurately defined.

Contrary to the prior proposition, distinctive errors prevailing in GPT-3 have diminished in Spanish-BFF-2 since the model shift. As previously mentioned, GPT-3 exhibited a recurring issue where approximately 11 % of definitions began with "A [lemma] is...,contravening the principle that the defined lemma should not feature in the description. However, in Spanish-BFF-2, less than 0,5 % of lemmas manifest this handicap, highlighting the improved lexicographic quality of GPT-4-turbo definitions. Finally, GPT-3 displayed additional errors, such as definitions composed in English and nouns defined as third-person forms of verbs, which have also been solved in Spanish-BFF-2.

## 7 Conclusions

Here, we introduce the second iteration of constructing a Spanish dictionary using GPT models. We suggest a model demonstrating superior performance and delivering enhanced linguistic information compared to its predecessor. The outcome yielded a more robust and comprehensive Spanish dictionary, as qualitative and quantitative analysis proved. Notably, GPT-4-turbo demonstrated a noteworthy capacity to circumvent hallucinations, a predominant burden in contemporary NLP systems. Nevertheless, the possible enhancement of GPT models in this field lies in their limited generation of polysemous words.

## 8 Limitations and future work

This approach is based on a list of lemmas, so we understand it is limited to languages with substantial resources, such as Spanish. Another option could be using a corpus and a lemmatizer, but we should note that not all languages have these resources. Using LLMs to generate the initial list of lemmas instead of listing former dictionaries or corpora might also be interesting.

In the future, we would also bear to exploit the nature of words whose definitions GPT does not know. Seemingly, they are rare words, but a sweeping scrutiny would be advised.

## 9 Ethics statement

We understand LLMs' possibilities for industry and future academic research. We intend to contribute to a better understanding and development of NLP and promote responsible use.

# A  Appendix 1. Adequate definitions

## A.1  Nouns

- limitación: Acción y efecto de limitar o limitarse.
- redondilla: Estrofa de poesía de cuatro versos octosílabos con rima ABBA.

## A.2  Verbs

- limitar: Poner límites o fronteras a algo.
- reducir: Hacer menor el tamaño, la cantidad, la intensidad o la importancia de algo.

## A.3  Adjectives

- limitable: Que se puede limitar o restringir.
- mímico: Perteneciente o relativo al arte de imitar o representar mediante gestos.

### A.4 Adverbs
- limitadamente: De una manera limitada o con restricciones.
- aquí: En este lugar o cerca de él.

### A.5 Examples
- simbolismo: El simbolismo fue una reacción contra el realismo del siglo XIX.
- fenomenal: El concierto fue fenomenal, no olvidaré esa noche.
- fenomenológico: El análisis fenomenológico busca describir la experiencia tal como es vivida.
- alabar: Siempre alaba las habilidades de su equipo.

### A.6 Adverbs cosine
- inexplicablemente. DLE: de manera inexplicable; BFF-1: de manera inexplicable; BFF-2: de manera inexplicable, que no se puede explicar o entender.
- parcamente. DLE: de manera parca; BFF-1: de forma parca; BFF-2: de manera escasa o limitada.

## B Appendix 2. Error analysis

### B.1 Words without GPT-4-turbo known definition
- jaharrar: Verbo: Desconocido, no tengo información para generar una definición o ejemplo.
- juagar: Nombre masculino: Parece ser un error tipográfico o una palabra inexistente en español.
- lemanaje: Nombre masculino: Término sin definición conocida, posiblemente un error o una palabra inexistente.

### B.2 DLE employing sinonyms
- carduzar: DLE: cardar.; Spanish-BFF-2: Acción de desgarrar o deshilachar tejidos o ropas.
- desatapar: DLE: destapar.; Spanish-BFF-2: Quitar o remover la tapa o lo que obstruye algo.

### B.3 Over-derivation hallucinations
- copear: Verbo. Pescar con cope. *Salieron a copear en la bahía durante la marea alta.*
- ollera: Nombre femenino. lugar donde se fabrican o venden ollas.
- convicio: Nombre masculino. Vicio o defecto. *El convicio de la estructura comprometía la seguridad del edificio.*
- copayero: Compañero de parranda o de copas. *Juan es mi copayero de cada fin de semana.*
- textal: Nombre masculina. Conjunto de textos, especialmente los que se refieren a un mismo tema o autor

### B.4 Over-correction hallucinations
- destace: Nombre masculino. Acción y efecto de destacar o sobresalir. El destace de su obra se dio en la última década.
- procinto: Nombre masculino. Estado inmediato y próximo de ejecutarse algo y, especialmente, en la milicia, el hecho de estar para darse una batalla. Cinta o tira de tela que se utiliza para atar o ceñir algo.

### B.5 Common names as proper names
- simón: Nombre masculino: Nombre propio de persona. *Simón es mi mejor amigo.*
- abitón: En la mitología griega, uno de los gigantes hijos de Gea, la Tierra, y Urano, el Cielo. *Abitón fue derrotado por los dioses olímpicos en la gigantomaquia.*

### B.6 Fabricated polysemy
- asaltador. DLE: que asalta.; BFF-2: que asalta.; persona que asalta.
- atropellado. DLE: que habla u obra con precipitación.; BFF-2: que ha sido objeto de un atropello.; que se hace de manera precipitada, sin orden ni concierto.
- baboseo. DLE: acción y efecto de babosear.; BFF-2: Acto de babosear o decir tonterías.; Exceso de baba o saliva.